\documentclass[conference]{IEEEtran}
\usepackage{cite}

\usepackage{hyperref}
\usepackage{amsmath}
\usepackage{amssymb}
\usepackage{bbm}
\usepackage{bm}
\usepackage{amsfonts}
\usepackage{algorithmic}
\usepackage{textcomp}
\usepackage[english]{babel}
\usepackage{subcaption}
\usepackage{graphicx}
\usepackage{microtype}
\usepackage{mathtools}
\usepackage{amsthm}
\usepackage{multirow}
\usepackage{pifont}
\usepackage{booktabs} 
\usepackage{xcolor}
\usepackage{tabularx}
\usepackage[
n,
advantage,operators,sets,adversary,landau,probability,
notions,logic,ff,mm,primitives,events,complexity,
oracles,asymptotics,keys]{cryptocode}
\usepackage{tikz}
\usetikzlibrary{matrix,shapes,arrows,positioning,chains, calc, fit,backgrounds,matrix,patterns}

\theoremstyle{plain}
\newtheorem{theorem}{Theorem}[section]

\theoremstyle{definition}
\newtheorem{definition}[theorem]{Definition}

\theoremstyle{remark}

\newcommand{\honestUsers}{\mathcal{V}_\mathcal{H}}
\newcommand{\adver}{\mathtt{b}}

\newcommand{\userD}{\mathtt{D}}
\newcommand{\userC}{\mathtt{C}}
\newcommand{\userL}{\mathtt{L}}
\newcommand{\userP}{\mathtt{P}}

\ifCLASSINFOpdf
\else
\fi
\begin{document}
%
\title{On the Conflict between Robustness and Learning in Collaborative Machine Learning}

\author{\IEEEauthorblockN{Mathilde Raynal}
\IEEEauthorblockA{SPRING Lab, EPFL\\
Lausanne, Switzerland}
\and
\IEEEauthorblockN{Carmela Troncoso}
\IEEEauthorblockA{SPRING Lab, EPFL\\
Lausanne, Switzerland}}

%

\IEEEoverridecommandlockouts
\makeatletter\def\@IEEEpubidpullup{6.5\baselineskip}\makeatother

\maketitle

\begin{abstract}
Collaborative Machine Learning (CML) allows participants to jointly train a machine learning model while keeping their training data private.
In many scenarios where CML is seen as the solution to privacy issues, such as health-related applications, safety is also a primary concern. 
To ensure that CML processes produce models that output correct and reliable decisions \emph{even in the presence of potentially untrusted participants}, researchers propose to use \textit{robust aggregators} to filter out malicious contributions that negatively influence the training process.
In this work, we formalize the two prevalent forms of robust aggregators in the literature. We then show that neither can provide the intended protection: either they use distance-based metrics that cannot reliably identify malicious inputs to training; or use metrics based on the behavior of the loss function which create a conflict with the ability of CML participants to learn, i.e., they cannot eliminate the risk of compromise without preventing learning.
\end{abstract}


%

\section{Introduction}
\label{sec:intro}

Collaborative Machine Learning (CML) allows users to jointly train a machine learning model. 
The main argument to favor CML instead of centralized learning techniques is \emph{privacy}: CML enables training of models without the users' data leaving their device. 
In CML, users share model \textit{updates} that act as learning proxies, e.g., gradients or model weights~\cite{fl}. 
CML architectures differ in the means they orchestrate the sharing of updates amongst participants, with the main two approaches being server-aided (e.g., Federated Learning~\cite{fl}), and peer-to-peer (e.g., Decentralized Learning~\cite{dl}).


In many applications where the privacy of CML is typically considered an advantage, \emph{robustness} is a critical property --i.e., the reliability and correctness of the model output (e.g., prediction or classification result). For example in health applications~\cite{health1, health2, health3, health4, health5, health6, health7, helath8, health9, health10}, where incorrect diagnosis can threaten the life of patient; or in autonomous driving~\cite{fldriving, FLdeepdriving, feddrive, driving4, driving5, driving6, driving7} where wrong predictions threaten the safety of passengers, bystanders, and the environment.
Even in fields where robustness is not essential for safety, it may be needed for economic reasons, such as avoiding manipulations that change content visibility in advertisement applications~\cite{ads} or change credit risk prediction~\cite{credit}.

Lack of robustness may stem from either non-malicious failures, such as users having low-quality data or an unreliable network connection that prevents them from participating in the network; or from attacks performed by malicious participants whose goal is to influence the behaviour of the trained model(s) in some undesirable way. There is a number of CML-tailored sophisticated functions to improve robustness~\cite{rfa, trimmedmedian, tm2, scclipping, ogclipping, krum, bridge, bylan, order, cao, cert, romoa, driving5, mixing, byrdie, xie, ubar, localpoison, fltrust, svmdl, vote, rofl} -- referred to as \textit{robust aggregators}~\cite{scclipping}, \textit{byzantine resilient defenses}~\cite{byrdie}, \textit{scoring mechanism}~\cite{krum}, or even \textit{poisoning detectors}~\cite{local}.
These functions enable participants to \textit{evaluate} the model updates they receive, and reject those that are deemed detrimental to the final model, i.e., participants learn as if there were only honest users in the network.

The privacy of CML has been studied in depth~\cite{privacyDL}, and there is a growing amount of attacks against state-of-the-art robust aggregators~\cite{local, scclipping, alie, imp, history, limits} and new robust aggregators protecting against state-of-the-art attacks~\cite{scclipping, local, tm2, ogclipping, history}. Yet, there is no formal study of robustness in collaborative scenarios. Existing work focuses solely in proving the efficiency of state-of-the-art robust aggregators against a selection of existing attacks, and does not discuss the effectiveness of these defenses against an strategic adversary that adapts to the detection algorithm. This follows the typical trend in systems' evaluation in which researchers iterate attacks and defenses to arrive at the state-of-the-art, such as in the search for robust algorithms in the centralized machine-learning literature~\cite{carlini}.
The lack of formalism and evaluation is even more problematic in light of the recent findings which demonstrate how standard experimental setups can misrepresent robustness~\cite{pitfalls}. 

In this paper, we provide the \emph{first} formalization of the robustness problem in CML and use this characterization to be the first to provide insights on the limits of the most prevalent robustness approaches in CML: in both approaches, for honest participants to learn they must always incur a risk of being manipulated. Such risk might be acceptable in some applications, like in advertising; or inadmissible in others, like in health or self-driving applications due to potential harms to users.
Concretely, we make the following contributions:
\begin{itemize}
    \item We systematize the different kinds of robust aggregation techniques in the CML literature. We find that most robust aggregators belong to one of two categories: those that evaluate the robustness of updates based on their distance, e.g., L2 distance to a reference point; and those that evaluate the robustness of updates based on their behavior, e.g., error rate.
    \item We provide a game-based formalization of the learning and robustness goals in CML. This formalization enables to reason in a principled way about the adequacy of robust aggregators design approaches and the trade-offs between these two goals.
    \item We demonstrate that distance-based robust aggregators that allow participants to learn from others must allow for adversarial manipulations of the same order of magnitude. This is because \emph{distance is not an adequate proxy for evaluating the maliciousness of updates}.
    \item We prove that when using behavior-based robust aggregators, the quality of the evaluation that users make is inversely proportional to their ability to learn from collaboration. Thus, the only users that can make good decisions and accurately reject malicious updates are those whose prior knowledge implies that they do not benefit much from participating in collaborative learning. Conversely, users who have the highest learning potential cannot produce accurate robustness evaluations, and remain vulnerable to manipulation despite their use of robust aggregation. 
    \item We empirically demonstrate the trade-off for both kinds of robust aggregators on state of the art aggregators, targeting fields where CML has been hailed as the solution to privacy.
    \end{itemize}
\section{Robustness in Collaborative Learning}

\noindent\textbf{Collaborative learning.} 
A Collaborative Machine Learning (CML) system consists of $n$ users who hold private local data. User $i$ holds data $S_i$ formed by tuples of samples $x$ and labels $y$ sampled from the domain $\mathcal{D} = (\mathcal{X}, \mathcal{Y})$.
Users are connected according to a graph $\mathcal{G}$, in which nodes are users and edges represent connections between users. We call this graph \emph{topology}. A common topology in CML is a star, used in Federated Learning~\cite{fl}, in which all nodes communicate with each other via a central entity.

In CML, users initialize their models with common public parameters~$m^0$ and iterate over the following three steps until a stop condition is met:
(1) \textit{Local training:} At the beginning of epoch $t$, users apply gradient descent on their local model parameters~$m^t$ using their local data $S_i$, and compute an \textit{update}~$\theta_{i}^{t}$.
(2) \textit{Communication:} Users share their model update $\theta_{i}^{t}$ with their neighbors--- or the server in the case of federated learning.
(3) \textit{Aggregation:} Users (or the server in federated learning) aggregate the updates, e.g., compute their average or their median. Users obtain an updated model $m^{t+1}$ using the result of their local aggregation. In federated learning, the server updates the global model and sends it to the users.


\smallskip\noindent\textbf{Robustness Adversary: Goal and Capabilities.}
The primary goal of CML is to enable users to obtain a model which performs better on the learning task than a model trained solely on their local data. In many cases, the application also requires this model to be \textit{robust}, i.e., the participants must not be able to influence the model in any undesired way. 
While \textit{undesired} is subjective and potentially relative to the application, in all cases the goal is to prevent manipulations that result in erroneous model outputs. 

We evaluate robustness when one, or more, of the participants are adversarial. We only consider as potential adversaries participants that send updates to other participants performing (robust) aggregation, e.g., to the federated server~\cite{fl} or to peers in decentralized learning~\cite{dl}. 
This excludes the scenario where a malicious federated server sends updates to users that accept it without further checks, as in that case, the server can already perform any adversarial manipulation by design.

The adversary aims to degrade the performance of the system, either in an \textit{untargeted} way, e.g., reducing the performance of the model as in a denial of service attack; or in a \textit{targeted} fashion, e.g., changing the output of some chosen inputs.
Following the literature, we assume this adversary to be malicious, i.e., they can send arbitrary model updates to other participants during training e.g.,~\cite[Section 3.2]{rfa}, ~\cite[Section 2]{krum}~\cite[Section 1]{scclipping},~\cite[Section 2.1]{rofl},~\cite[Section 1]{trimmedmedian},~\cite[Section 3.C]{baffle}.

The precise capabilities of the adversary depend on the CML protocol and the threat model of the application~\cite{open}.
At a minimum, the adversary, who participates in the training, has knowledge of the training public parameters (model architecture, learning rate) and of the updates it receives from its neighbors. 
Other capabilities such as the ability to participate in all epochs can vary depending on the CML system.

\section{Systematization of Robust Aggregators}

In CML, the main tool to ensure robustness of the final model is the use of \textit{robust aggregators} during the training process. Robust aggregators aim to filter out detrimental updates before  aggregation, such that only honest users contribute to training.
Robust aggregators rely on an evaluation metric to distinguish benign updates from malicious ones.
We use $R(\theta; \mathcal{D})$ to refer to the evaluation of the update $\theta$ on the inputs from $\mathcal{D}$, using the metric $R$.
The output of the evaluation is used in conjunction with a threshold value $\delta$ to identify benign updates, e.g., $R(\theta; \mathcal{D})<\delta$.

A robust aggregator is a concrete instantiation of the metric $R$ and a means to \textit{evaluate} the exchanged updates, i.e., each proposition of a metric $R$ is a proposition of a definition of robustness.
The metric $R$ must have two properties.
First, $R$ must capture the difference between benign and malicious updates such that the robust aggregator can effectively reject malicious updates: if an update does not influence the model in any way that leads to erroneous results, then $R(\theta; \mathcal{D})$ must be below $\delta$.
Second, participants must be able to correctly compute $R$, even when they have access to a limited dataset and limited knowledge on the learning task.


\smallskip\noindent\textbf{Robustness metrics.} The metric $R$ should, ideally, be informative about the maximal error of the model $m^{t+1}$ resulting of the use of the evaluated update $\theta$ on the current model $m^t$ on inputs from the domain $\mathcal{D}$, i.e., $R(\theta, \mathcal{D}) \doteq \underset{(x,y)\sim\mathcal{D}}{max}[m^{t+1}(x) - y]$.
Yet, computing such metric in the collaborative setting is impossible because it requires computing the output of the model on \textit{all} samples from $\mathcal{D}$.
In addition to being a potentially unbounded computational task, individual users do not know the whole domain of inputs (by definition of the problem).


As a consequence, metrics $R(\theta, \mathcal{D})$ proposed in the literature resort to the evaluation of \textit{proxies} that can be easily computed.
These proxies include (but are not limited to): average accuracy of the update on a fixed test set~\cite{xie}, invariance of prediction under input perturbation~\cite{invariance}, smoothness of decision regions~\cite{advex} when the update is a model itself, or more generally the low norm of weights~\cite{l2} or L2-distance to the average of previously received updates~\cite{trilemma}.

In general, the evaluation of an update $\theta$ is based on a comparison between this update and a reference that we denote as $\theta^*$.
The reference can be a gradient or a model, typically of the same format as $\theta$. 
In the literature, we find that the reference $\theta^*$ is either fixed before training given prior knowledge of the participants, or computed dynamically per epoch given information available at the time of evaluation. In the latter case, the evaluator can use local knowledge (e.g., a local update), or knowledge obtained from the other participants' in the network (e.g., the average of all updates, assuming that the majority of users is honest).  
The closer the update being evaluated ($\theta$) is to this reference ($\theta^*$), according to $R$, the more likely $\theta$ is to be benign.
When $\theta=\theta^*$, $R(\theta, \mathcal{D})=0$.

\smallskip\noindent\textbf{Robustness metrics in the literature.} We find two prevalent classes of metrics in the CML literature to assess `closeness', which we formalize in the next two sections.
Systematizing robust aggregators through classes of metrics allows us to reason about the properties of a class instead of individual robust aggregators. It also ensures that our results apply to new aggregators without the need of an analysis from scratch.
We discuss classes of metrics not covered by our analysis in Section~\ref{sec:outofscope}.



\subsection{Distance-based Evaluation}

The most common class of robust aggregators are those that evaluate a characteristic of the update itself, e.g., the update weights ~\cite{rfa, trimmedmedian, tm2, scclipping, ogclipping, krum, bridge, bylan, order, cao, cert, romoa, driving5, mixing, byrdie, rofl, fltrust, ubar, sentinel}.
Approaches in this class measure the closeness between the update $\theta$ and a reference $\theta^*$ using a \textit{distance metric} $\Delta$ ---i.e. a metric that has reflexivity, non-negativity, symmetry, and respects the triangle inequality.
We refer to this class of robust aggregators as \textit{distance-based}.
We formalize them in Definition~\ref{def:distance_agg} and give examples of possible $\theta^*$ and $\Delta$ in Section~\ref{sec:exsdist}.

\definition[Distance-based robustness]~\label{def:distance_agg}
We say a robust aggregator is distance-based if its evaluation function can be expressed as:
$$R(\theta, \mathcal{D}) \doteq \Delta(\theta, \theta^*)$$
for a distance metric $\Delta$ and a reference update $\theta^*$.

\subsubsection{Examples of distance-based robust aggregators}\label{sec:exsdist}\

\noindent\textbf{Multi-KRUM~\cite{krum}}. 
Multi-KRUM is a robust Federated Learning aggregator function run by the server. It uses the KRUM score as the robustness evaluation function $R$. 
Multi-KRUM is an example of an aggregator whose reference point $\theta^*$ \emph{dynamically} computed at each epoch using \emph{network knowledge}. 
The KRUM score of an update $\theta$ captures the cumulative L2-distance between this update and the users' updates.
The server computes $R(\theta, \mathcal{D}) = score(\theta) =  \sum_{\theta_j\in CU(\theta)} ||\theta - \theta_j||_2,$
where $CU(\theta)$ are the $n-f-2$ closest updates (CU) to $\theta$ from all received updates.
The KRUM score can be expressed as $\Delta(\theta, \theta^*)$ with $\Delta$ the L2-distance, and $\theta^*$ a function of users' updates such that $\sum_{\theta_j\in CU(\theta)} ||\theta - \theta_j||_2 = ||\theta - \theta^*||_2$.
The Multi-KRUM robust aggregator selects the $n-f$ updates with lowest score, and outputs their average as result of the aggregation.

\smallskip\noindent\textbf{RoFL~\cite{rofl}}.
ROFL is another robust Federated Learning aggregator function run at the server. It defines the evaluation function $R$ to be the norm of the update. 
ROFL is an example of an aggregator that uses a \emph{fixed} reference point $\theta^*$ \emph{decided a-priori}. 
In RoFL, users construct cryptographic proofs that the norm of their update is below a threshold selected by the server.
Encryption aside, the RoFL aggregation function can be instantiated with $\Delta$ the L2-distance, and $\theta^*$ the all 0-model, i.e., $R(\theta, \mathcal{D}) = ||\theta||_2$.

\smallskip\noindent\textbf{Self-Centered Clipping~\cite{scclipping}}.
Self Centered Clipping (SCC) is a peer-to-peer-specific robust aggregation scheme. It uses the L2-distance as evaluation function $R$, and is an example of an aggregator whose reference point is \emph{dynamically computed each epoch based on local knowledge}.
In SCC, users evaluate an update $\theta$ by computing the L2-distance between this update and their own local update $\theta_i$:
$R(\theta, \mathcal{D}) = ||\theta - \theta_i||_2.$
This function can be expressed as $\Delta(\theta, \theta^*)$ with $\Delta$ the L2-distance, and $\theta^*$ the user's own update $\theta_i$.
SCC differs from other evaluators in that it does not directly reject the updates for which $R(\theta, \mathcal{D}) > \delta$, but clips them into an update $\theta' = \texttt{min}(1, \delta /||\theta - \theta_i||)\cdot (\theta - \theta_i) + \theta_i$ that satisfies $R(\theta', \mathcal{D}) \leq \delta$, and discards the rest of the update.
The output of the aggregation is the average of the accepted (clipped) updates.

\subsection{Behavior-based Evaluation}

A second common class of robust aggregators are those that evaluate the behavior of the model after integrating the received update (or the behavior of the update itself when directly evaluated on the model) on selected inputs~\cite{xie, localpoison, svmdl, vote, basil, baffle, bygars, ubar, sentinel}.
These evaluators compare this behavior to a pre-determined expected behavior modeled by a function $\pi$, e.g., the expected loss of the model.
The reference $\theta^*$ determines the expected behavior. $\theta^*$ itself does not necessarily need to be known. Sometimes only its behavior $\pi(\theta^*, \mathcal{D})$ matters for the evaluation purpose. 
We refer to this category of robust aggregators as \textit{behavior-based}.
We formalize them in Definition~\ref{def:behavior} and give examples of choice of $\theta^*$ and $\pi$ in Section~\ref{sec:examples-behavior}.

\definition[Behavior-based robustness]~\label{def:behavior}
We say a robust aggregator is behavior-based if its evaluation function can be expressed as: $$R(\theta, \mathcal{D}) \doteq \pi(\theta; \mathcal{D}) - \pi(\theta^*; \mathcal{D})$$
for a given function $\pi$ deterministic in $\mathcal{D}$ and reference $\theta^*$.


\subsubsection{Examples of behavior-based robust aggregators}\label{sec:examples-behavior}\

\noindent In our CML literature review, we have not found an example of behavior-based robust aggregator which uses a fixed reference point $\theta^*$ nor fixed behavior $\pi(\theta^*, \mathcal{D})$. Thus, we do not include examples of this category in the paper. We note that our evaluation methods seemingly apply to this case. Examples of other categories follow.

\smallskip\noindent\textbf{RD-SVM~\cite{svmdl}}. 
RD-SVM is an example of an aggregator updated \emph{dynamically per epoch based on local knowledge}.
Users in RD-SVM evaluate updates $\theta$, which are model weights, by comparing these updates' expected hinge loss~\cite{hingeloss} to the hinge loss of their current local model $m_i^t$. 
The hinge loss encourages correct predictions with a margin, penalizing misclassifications more severely than traditional classification loss functions like cross-entropy.
In RD-SVM, $R(\theta, \mathcal{D})$ can be expressed as $\pi(\theta, \mathcal{D}) - \pi(\theta^*, \mathcal{D})$ with $\pi$ the hinge loss on a test set sampled from $\mathcal{D}$, and $\theta^*$ the user's current model $m_i^t$.
RD-SVM is an example of an aggregator whose reference point is
computed on \emph{local knowledge.}
Users reject updates $\theta$ with a higher hinge loss on the test set ($R(\theta, \mathcal{D}) > 0$). 

\smallskip\noindent\textbf{Error Rate based Rejection~\cite{localpoison}}. 
ERR is an example of an aggregator updated \emph{dynamically per epoch based on network knowledge}.
In ERR, users compute two models. First, a model, $m$, where the received update is aggregated into the current model together with other available updates. Second, a model $\theta^*$ where only the other updates are aggregated into the current model. 
$R(\theta, \mathcal{D})$ is then computed as $\pi(\theta, \mathcal{D}) - \pi(\theta^*, \mathcal{D})$, with $\pi$ the error rate on a test set sampled from $\mathcal{D}$ and $\theta^*$ the model computed using other users' model update, i.e., using \textit{network knowledge}.
In ERR, users reject the $f$ received updates with largest $R(\theta, \mathcal{D})$, i.e., those that increase the error rate the most. 



\subsection{Out of Scope Robust Aggregators}
\label{sec:outofscope}
In this work, we do not study data-driven approaches~\cite{clean}, since in CML, the data of users is not available to protect privacy. 
We also do not study techniques which assume that participants --including adversaries-- behave honestly, as adversaries aiming to influence the model are malicious by definition.
This includes all approaches where users enable robustness by applying robust algorithm(s) locally such as data cleaning~\cite{clean, demon, fail, roni} or use of robust loss function to perform training~\cite{flamingo, robloss1, robloss2, robloss3, local, naive, cert, AFL, afine}.
Finally, we do not study robust aggregators that are designed against non-adversarial failures, such as reliability of communications~\cite{reliable} or resilience to user dropout~\cite{flamingo}.

Our two categories cover all other techniques used by robust aggregators that we are aware of in the literature.

\section{CML goals: Robustness and Learning}

To enable the study of robustness and its implications beyond empirical evaluation of attacks, we design a formal game-based framework to capture the two properties of collaborative learning we are concerned with in this paper: the ability to learn, and the ability to do so without risk of interference by malicious participants in the collaborative learning process.

\subsection{Robustness Indistinguishability}

\begin{figure}[!h]

    \begin{tikzpicture}
    \matrix (m)[matrix of nodes, column sep=-\pgflinewidth, name=m, nodes={draw=none, anchor=west} ]{
    \phantom{.} & \phantom{.} & \phantom{.} \\
    \phantom{.} & \phantom{.} & \phantom{.} \\
    Distinguisher $\userD$ & & Challenger $\userC$\\
    & & $b \leftarrow\$ \{0, 1\}$ \\
    & & $\theta_0 \leftarrow \mathcal{O}_\texttt{benign}(\mathcal{D}, \delta)$\\
    & & $\theta_1 \leftarrow \mathcal{O}_\texttt{malicious}(\mathcal{D}, \delta)$\\
    & $\theta_b$ & \\
    $b’ \leftarrow \userD (\theta_b, S_\userD, \mathcal{O}_\texttt{auxiliary})$ & & \\
    $\userD$ wins if $b’ == b$ & & \\
    };
    \node[fit=(m-1-1)(m-1-3), draw=none]{$\text{R-IND}_\userD(\delta; S_\userD, \mathcal{D}, \mathcal{O}_\texttt{benign}, \mathcal{O}_\texttt{malicious}, \mathcal{O}_\texttt{auxiliary}):$};
    \draw[shorten >=-3.cm] (m-1-1.south west)--(m-1-3.south east);
    \draw[shorten <=-.5cm] (m-3-1.south east)--(m-3-1.south west);
    \draw[shorten <=-.5cm] (m-3-3.south east)--(m-3-3.south west);
    \draw[shorten <=-.3cm,shorten >=-.3cm,-latex] (m-7-2.south east)--(m-7-2.south west);
    \end{tikzpicture}

\caption{R-IND Game}
\label{fig:rind}
\end{figure}
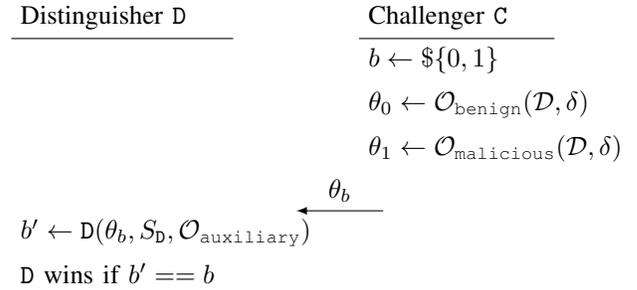

We first design a robustness-indistinguishability game (see Figure~\ref{fig:rind}) that captures whether a distinguisher, i.e., a participant, can use a function $R$ to effectively distinguish benign updates from malicious updates, and whether this distinguisher can compute this function $R$ using solely its local knowledge.


During the robustness indistinguishability game, the challenger generates two updates, one benign and one malicious, with the help of two oracles: $\mathcal{O}_\texttt{benign}$ and $\mathcal{O}_\texttt{malicious}$, and presents one of them to the distinguisher.
Given a dataset $S_\adv$ sampled from the domain $\mathcal{D}$ and access to an oracle $\mathcal{O}_\texttt{auxiliary}$, the task of the distinguisher is to distinguish whether the update it receives is benign or not.
The $\mathcal{O}_\texttt{auxiliary}$ oracle models any auxiliary information on the updates available to the distinguisher.
If the distinguisher wins the R-IND game, it means that they are able to distinguish benign from malicious updates using the information available to them.

\smallskip\noindent\textbf{Instantiating the R-IND game.}
The mapping from the R-IND game to the context of CML is quite straightforward: the distinguisher in the game represents an honest participant running a robust aggregation protocol to evaluate the updates they receive with the goal of filtering those that would affect their model's performance.
The inputs $\mathcal{D}$ and $S_\adv$ represent the input domain and the local data of the said participants, respectively. 
The oracle $\mathcal{O}_\texttt{auxiliary}$ captures the information available to the participant other than their local training data, if any.
For example, this oracle can model the ability of the server to observe the updates produced by all users before performing the evaluation of individual updates.
The challenger represents other users of the network, whose benignness is not guaranteed.
The oracle $\mathcal{O}_\texttt{benign}$ generates updates produced by honest users, i.e., if the challenger is benign and the oracle $\mathcal{O}_\texttt{malicious}$ generates updates produced by malicious users.
Using oracles allows us to reason about \textit{any} strategic adversary beyond a precise attack. 

If the distinguisher can win this game, instantiated with specific inputs and oracles, and using the evaluation function of a chosen robust aggregator, it means that this aggregator enables participants to successfully evaluate the maliciousness of received updates.


\subsection{Users' Learning Potential}

\begin{figure}[!h]

    \begin{tikzpicture}
    \matrix (m)[matrix of nodes, column sep=-\pgflinewidth, name=m, nodes={draw=none, anchor=west} ]{
    \phantom{.} & \phantom{.} & \phantom{.} \\
    \phantom{.} & \phantom{.} & \phantom{.} \\
    Learner $\userL$& & Challenger $\userC$\\
    & & $\mathcal{L}^* = \underset{m^*}{\text{ min }}\underset{(x, y)\sim\mathcal{D}}{\mathbb{E}}[\mathcal{L}(m^*, (x, y))]$ \\
    $m \gets \userL (S_\userL)$& & \\
    &\phantom{.} $m$ \phantom{.}&\\
    & & $\mathcal{L}_\userL = \underset{(x, y)\sim\mathcal{D}}{\mathbb{E}}[\mathcal{L}(m, (x, y)]$\\
    $\userL$ wins if & & \\
    $\mathcal{L}_\userL<\mathcal{L}^*+\epsilon$ & & \\
    };
    \node[fit=(m-1-1)(m-1-2), draw=none]{$\text{L-EUF}_\userL(\epsilon; S_\userL, \mathcal{D}):$};
    \draw[shorten >=-4.5cm] (m-1-1.south west)--(m-1-3.south east);
    \draw[shorten <=-.5cm] (m-3-1.south east)--(m-3-1.south west);
    \draw[shorten <=-.5cm] (m-3-3.south east)--(m-3-3.south west);
    \draw[shorten <=-.3cm,shorten >=-.3cm,-latex] (m-6-2.south west)--(m-6-2.south east);
    \end{tikzpicture}

\caption{L-EUF Game}
\label{fig:leuf}
\end{figure}
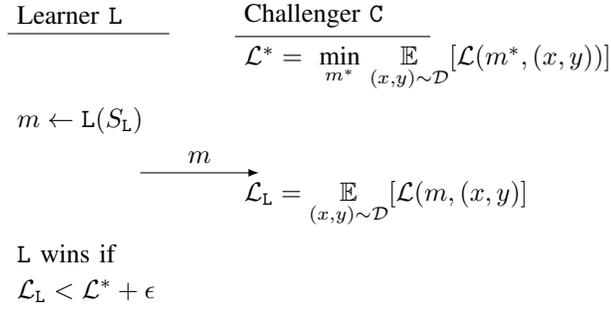

We also design a forgery game that captures the ability of users to learn from their peers in the collaborative learning process.
The goal of the participant playing the game, the learner, is to forge a model which performs well for a target task, given the data they have knowledge of.
In the learning existential (un)forgeability game (L-EUF, see Figure~\ref{fig:leuf}), the learner is given a dataset $S_\userL$ sampled from the domain $\mathcal{D}$. Then, a challenger challenges them to forge (i.e., produce) a model with a loss $\epsilon$-close to the one of a model trained on the whole domain $\mathcal{D}$ distribution, i.e., the optimal learner.


The advantage of the learner in winning the L-EUF game is tied to how well they are able to learn the target task given their local data.

\smallskip\noindent\textbf{Instantiating the L-EUF game.}
The learner in the game represents an honest participant with the goal to produce a model as good as possible on the learning task at hand.
Similarly to the R-IND game, the inputs $\mathcal{D}$ and $S_\userL$ represent the input domain and the local training set of the participants.
The loss function $\mathcal{L}$ and the architecture of the model $m$ are parameters of the training process, and thus we do not model them inside the game.
If the learner can win given specific inputs, it means that an user in CML can learn a $\epsilon$-approximately correct model on $\mathcal{D}$ using their local data $S_\userL$ only.
Finally, $\epsilon$ acts as a difficulty parameter.
The smaller $\epsilon$ is, the closer the forged model needs to be to the optimal learner.

The L-EUF game is a useful tool to capture the benefits of collaborative learning. We can do so by comparing the advantage of a  participant in winning the game when using its local data only ($S_\userL$), with the advantage when using the union of the data of all participants (simulating a centralized scenario where all data has been shared). The latter represents the best case scenario of collaboration: all users learn as if they had all data, i.e., as if there was neither noise nor loss of information due to computation, representation, or aggregation.

We call the difference between the advantage of games with these two inputs the \textbf{learning potential} of the learner, i.e., CML participant, given a CML network. The learning potential represents how much more a participant could learn if they would gain all knowledge from others.
For a participant with poor local data whose advantage in winning the game is small (in fact they cannot win), the difference with respect to the model trained on all data will be large indicating that they greatly benefit from collaborative learning. 
For a participant that can win the game on their own, and whose advantage is thus large, there will be not much difference with the model trained on all data, capturing that this participant has little to learn from others and thus does not have much incentive to collaborate.

Even though in this paper we mostly discuss the learning potential for participants holding local data being \textit{users}, the concept modeled by the L-EUF game also applies to a \textit{server} in the federated setting. While the server does not participate in CML to learn a model for itself, it learns a model \textit{on behalf} of the users.
The server's learning potential is then a metric of how much the server needs inputs from the users to perform well at the learning task.
If the server has a small learning potential, the server has very little to learn from users' updates. 
In such case, the best strategy for the users would be to let the server rely on centralized training where the server would compute a model on its own local data and broadcast it to the network.

\section{Distance is a bad Proxy for Maliciousness}

We now study whether distance-based aggregators that use evaluation functions $R(\theta, \mathcal{D})$ of the form $\Delta(\theta, \theta^*)$ enable CML participants to accurately distinguish benign from malicious updates. We also characterize the relationship between the robustness and learning properties under this robust aggregation approach.

\subsection{Space for Learning is space for Manipulation}
\subsubsection{Distance to win the R-IND game}\

\noindent During the R-IND game, the distinguisher classifies updates $\theta_b$ as benign (outputs $b'=0$), if the distance from this update to the reference point $\theta^*$ is below a given threshold $\delta$, according to the metric $\Delta$. Otherwise, it classifies the update as malicious (outputs $b'=1$).
If there exist adversarial updates whose distance to $\theta^*$ is below $\delta$, then there exists an instantiation of the oracle $\mathcal{O}_\texttt{malicious}$ for which the distinguisher will fail to identify updates as malicious using a distance-based evaluation function.
In this scenario, the distinguisher cannot win the R-IND game using $R=\Delta(\theta_b, \theta^*)<\delta$.
In practice, this means that a strategic adversary can always construct malicious updates that a distance-based robust aggregator would classify as benign, and a user in CML would accept malicious updates during training.
It follows that, for a distance-based evaluation to provide robustness, the choice of $R$ and its parameters$\Delta$, $\theta^*$, and $\delta$, must guarantee that all updates $\theta$ such that $\Delta(\theta, \theta^*)<\delta$ are benign. It is also required that all updates $\theta$ such that $\Delta(\theta, \theta^*) > \delta$ are malicious, such that benign updates are not rejected and learning from collaboration is enabled.
In other words, $R$ must be chosen such that a participant $\userP$ using this metric wins the R-IND game, i.e., that $Adv_\userP^\text{R-IND}(\delta; S_\userP, \mathcal{D})=1$.

\begin{theorem}~\label{theorem:dist}
When $R(\theta, \mathcal{D})$ is distance-based, there exist a equivalence reduction from the R-IND game to the L-EUF game, meaning that if, for a participant $\userP$ and a threshold $\delta^*$, $Adv_\userP^\text{R-IND}(\delta^*; S_\userP, \mathcal{D})=1$, then $Adv_\userP^\text{L-EUF}(\epsilon; S_\userP, \mathcal{D})=1$ for all $\epsilon$. 
\end{theorem}

The implication of this theorem is that a participant $\userP$ which succeeds in designing $R$ that accurately distinguishes malicious updates from benign ones, also succeeds on their own at the learning task and does not benefit from collaboration.
As a result, there exist no participant which achieves robustness and has potential for learning at the same time when using distance-based evaluations.

\begin{proof}[Proof Sketch]
Let us assume that a participant can choose the parameters of $R$, $\Delta$ and $\theta^*$, and $\delta$ such that they perfectly distinguish benign and malicious updates, thus can win the R-IND game with advantage 1.
The knowledge needed to make this choice is the same knowledge required to build a distinguisher towards learning, i.e., to distinguish updates that are beneficial with regards to the learning task from the ones that are detrimental.
Furthermore, a participant able to distinguish good and bad updates with regards to learning is also able (up to computational hardness) to \textit{generate} good updates towards learning, by finding any update $\theta$ within distance $\delta$ of $\theta^*$ according to $\Delta$ that the learning distinguisher accepts.
With these updates, the participant can forge a model that will succeed at the learning task.
While this step is inefficient, it allows the participant to wins the L-EUF game for $\epsilon=0$ with advantage 1.
Since $Adv_\userP^\text{L-EUF}(0; S_\userP, \mathcal{D}) \leq Adv_\userP^\text{L-EUF}(\epsilon; S_\userP, \mathcal{D})$, i.e., the participant can use the output of L-EUF$(0; S_\userP, \mathcal{D})$ to win L-EUF$(\epsilon; S_\userP, \mathcal{D})$ for all $\epsilon > 0$, then the participant wins the L-EUF game for all $\epsilon$ with advantage 1.

\end{proof}

\subsubsection{Advantage to win R-IND Compose during Training}\

\noindent As we mention above, whenever $Adv_\userP^\text{R-IND}(\delta; S_\userP, \mathcal{D})$ is \emph{not} 1, and thus our theorem does not apply, there exist an adversary that can cause participants (distinguishers) to make errors, and therefore manipulate their models. 
This is the case for state-of-the-art distance-based robust aggregators, where $R$ is chosen to upper-bound the capability of the adversary to manipulate the model in each update, i.e., $R$ bounds the norm of updates~\cite{rofl}. 
Hence, these aggregators \textit{cannot} guarantee that accepted updates are benign inputs from honest users, i.e., they do not guarantee that $R$ consistently wins the R-IND game with advantage 1.
Making a parallel with gradient descent: controlling the size of the steps but not the direction is not a sound technique to reach a meaningful state---at least not without further assumptions.

The impact of the inability to accurately distinguish benign from malicious updates for one instance of the R-IND game is worsened by the fact that the CML training process is iterative. 
Hence, participants do not play one instance of a R-IND game, but a \emph{composition} of games corresponding to the sequential epochs of the training.
The adversary can leverage this situation by crafting updates that individually are closer to the reference than the threshold $\delta$, effectively enabling the victim participant to win all individual instances of R-IND games across epochs; but when the updates are considered together they have a magnitude larger than the threshold ($\delta\cdot \tau$ over $\tau$ epochs).

For example, assume an adversary that is interested in manipulating the current model $m$ to obtain a target $m_{target}$ (e.g.,a backdoored version of the model).
The use of a distance-based aggregator forces the adversary to only use updates for which  $\Delta(\theta, \theta^*) < \delta$. 
To achieve their goal, the adversary must influence the model by an amount $\Delta(m_{target}, m) = \delta_{target}$. Then, it suffices that the adversary creates $\tau \thickapprox \delta_{target} / \delta$ small combined updates to perform the desired manipulation.
Thus, the only question is how many epochs the adversary needs to achieve the goal. In general, the larger $\delta$, the less epochs the adversary needs to reach its goal.
It might be the case that the number of epochs of the training is fixed in advance, and lower than $\tau$.
In that case, the adversary can adjust its choice of $m_{target}$ to cause maximal security breach under both constrained step size and constrained number of epochs.

So, if learning potential is to be preserved, distance-based metrics cannot prevent the model from being manipulated. At most, they can increase the number of epochs the adversary needs to achieve their goal, as this number is proportional to $\delta$. Hardening the adversary's task, however, comes at a cost: when $\delta$ is small, the adversary requires many epochs to fulfill their goal, but a small $\delta$ also limits the magnitude of contributions of the honest users decreasing learning speed.
Increasing $\delta$ to admit (potentially) more informative benign updates and increase learning, also facilitates the task of the adversary that can manipulate the model in less rounds. Thus, when learning is enabled, i.e., $Adv_\userP^\text{R-IND}(\delta; S_\userP, \mathcal{D})$ is \emph{not} 1, \textbf{regardless of the metric $\Delta$ the acceptance threshold $\delta$ represents a trade-off between sacrificing learning and opening the door to attacks.}
\begin{figure*}[!ht]
    \centering
    \begin{subfigure}{0.45\textwidth}
    \includegraphics[width=\textwidth]{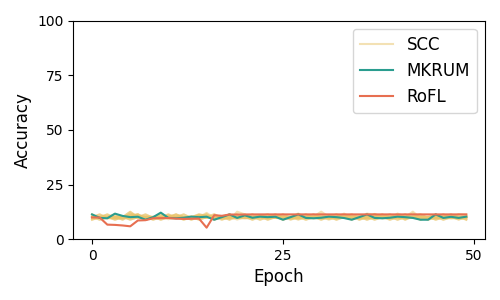}
    \end{subfigure}
    \begin{subfigure}{0.45\textwidth}
    \includegraphics[width=\textwidth]{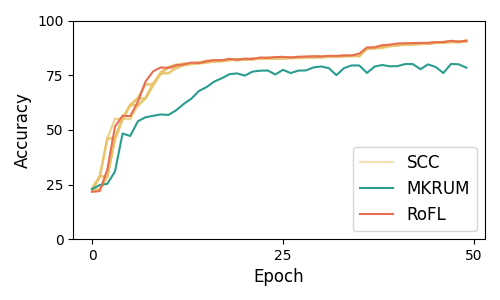}
    \end{subfigure}
    
    \begin{subfigure}{0.45\textwidth}
    \includegraphics[width=\textwidth]{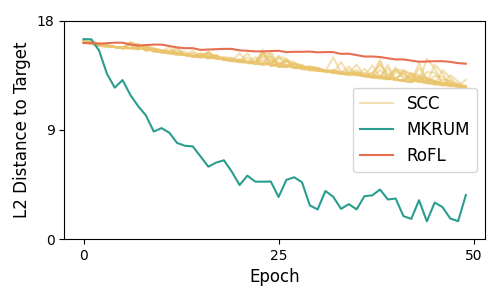}
    \end{subfigure}
    \begin{subfigure}{0.45\textwidth}
    \includegraphics[width=\textwidth]{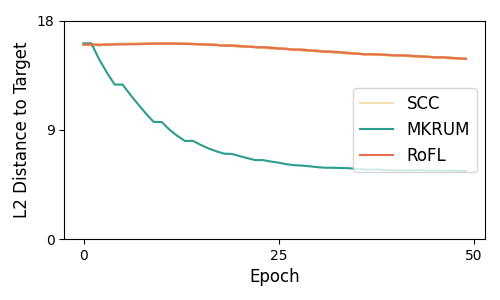}
    \end{subfigure}
    \caption{Accuracy (top) and L2 distance to adversarial target (bottom) under the state-override attack on the MNIST task (non-IID setup, left; IID setup, right).}
    \label{fig:mnist_iid}
\end{figure*}

\subsection{Validation on Existing Distance-Based Aggregators}\label{sec:eval-distance}
In this section, we conduct experiments to empirically support our theoretical findings.
We investigate the robustness of the training process for three distance-based robust aggregators in Section~\ref{sec:exsdist}: SCC~\cite{scclipping}, MKRUM~\cite{krum}, and RoFL~\cite{rofl}, in various CML tasks against a strategic adversary.

\subsubsection{Implementing a strategic adversary}\

\noindent We implement a strategic adversary by having the malicious participant(s) implement the state-override attack~\cite{privacyDL}.
In this attack, the adversary sends to each neighboring participant $i$ (either user or server) the following update: 

$$\theta_{\adv i} = n\cdot (\theta_{target} - \frac{1}{n}\sum_{j \neq \adv} \theta_j)\,,$$

which contains the negative of the sum of all other updates in the epoch.
As a result, when averaging the updates, the adversary's neighbors aggregation result is overwritten with $\theta_{target}$. In the formulation above, the success of the attack relies on the adversary having access to all neighbors' updates before sending its own. This can be a limitation if the system imposes a schedule. In that case, it is possible to use an approximation of the updates based on prior epochs with limited impact on the attack result~\cite[Appendix C]{privacyDL}.

The adversary can select a target $\theta_\texttt{target}$ to achieve various adversarial goals, e.g., controlling a few weights of $m^t_i$, e.g., inserting a backdoor~\cite{backdoor}, or to completely override $m^t_i$ with an arbitrary model. In our experiments, we assume the adversary's goal is to override the victims' model with the all-0 model.
We use the state-override attack as introduced in Pasquini et al.~\cite{privacyDL} when attacking MKRUM and RoFL. When attacking SCC, we refine the attack to be: 
$$\theta_{\adv i}^\text{SCC} = -\sum_{j \neq \adv} \texttt{min}(1, \delta /||\theta_j - \theta_i||)\cdot (\theta_j - \theta_i),$$ 
in order to cancel out the \textit{clipped} updates received by the neighbor $i$.



\subsubsection{Results on MNIST}\

\smallskip\noindent\textbf{Setup.}
For our experiments, we use the aggregators' default parameters values: for SCC, we set a clipping value $\delta = 1$ ($\delta$ is $\tau$ in the original paper); for MKRUM we set the threshold $\delta$ such that the number of accepted updates is $n - f=(n+3)/2$; and for RoFL  we set $\delta=1$ ($\delta$ is $B$ in the original paper, and we choose the value for which their results apply across datasets~\cite[Figures 4 and 5]{rofl}). 
For MKRUM and RoFL, designed for the federated learning setting, we use a standard star topology. For SCC, motivated by its use in concrete applications~\cite{disco} we consider a fully-connected topology.

We use the MNIST dataset in a typical setting: we divide the dataset into train and validation sets; a shallow 3-layers CNN model architecture; an Adam optimizer; and default hyper-parameters: batch size of 32, learning rate of 0.01; and train for 50 epochs.
We consider 20 users, 18 honest users and 2 adversaries. This ratio of malicious versus honest nodes (10\%) is small compared to the literature (e.g., in MKRUM the ratio is 45\%, and in SCC it is 22\%). 

\smallskip\noindent\textbf{Results.}
We first measure the impact of data distribution on the robustness of the training process.
We test users whose local datasets are IID and non-IID fashion.
In Figure~\ref{fig:mnist_iid}, we report the accuracy of honest users in different epochs for all robust aggregators (top); and the $\Delta$-distance (i.e., L2-distance for our tested aggregators) of the users' model to the target, i.e., $\Delta(m^t, m_\texttt{target})$  (bottom).
In the non-IID setting (left), no aggregator permits the users to learn (the accuracy does not increase). Yet, the attack is effective, and we see how the distance to the target model decreases. When using MKRUM, the distance approaches zero, indicating the adversary eventually takes control of the users' model.
In the IID setup (right), users do learn (accuracy increases), but the attacker still succeeds albeit slower than in the non-IID case.
We also observe a similar behavior of the SCC and RoFL aggregators (the lines overlap on the bottom right graph), which could indicate that they implement the same functionality under an IID setup and a complete topology in peer-to-peer. 
Overall, we show that none of the tested distance-based aggregators perfectly distinguish malicious from benign updates.

In summary, if users have heterogeneous non-IID data, the studied robust aggregators prevent all learning, and cannot prevent the attack.
When users have homogeneous data, users still learn under the attack, which is less effective, but we note that given the trend of the distance to the target model, the attack would eventually succeed.

\begin{figure*}[!htbp]
    \centering
    \begin{subfigure}{0.31\textwidth}
    \includegraphics[width=\textwidth]{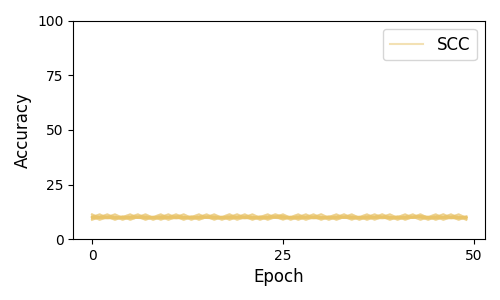}
    \end{subfigure}
    \hfill
    \begin{subfigure}{0.32\textwidth}
    \includegraphics[width=\textwidth]{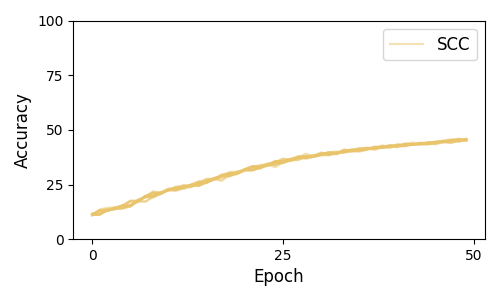}
    \end{subfigure}
    \hfill
    \begin{subfigure}{0.32\textwidth}
    \includegraphics[width=\textwidth]{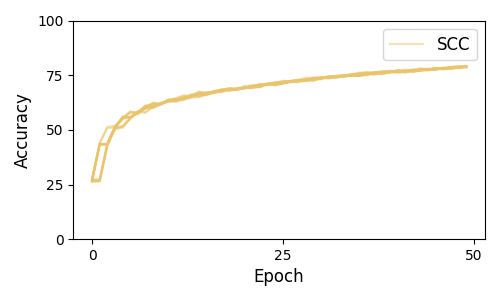}
    \end{subfigure}

    \begin{subfigure}{0.32\textwidth}
    \includegraphics[width=\textwidth]{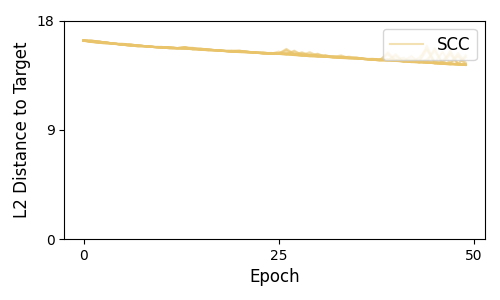}
    \end{subfigure}
    \hfill
    \begin{subfigure}{0.32\textwidth}
    \includegraphics[width=\textwidth]{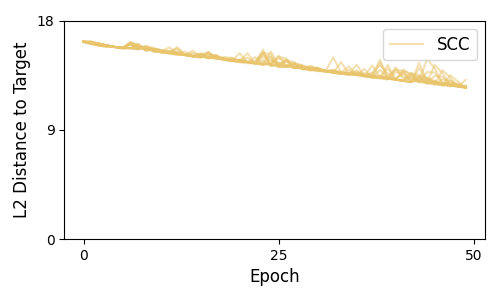}
    \end{subfigure}
    \hfill
    \begin{subfigure}{0.32\textwidth}
    \includegraphics[width=\textwidth]{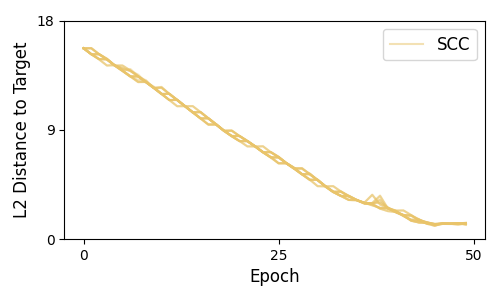}
    \end{subfigure}
    \caption{Accuracy during benign training (top) and L2 distance to the adversarial target under the state-override attack (bottom) on the MNIST task for $\delta=0.5$ (left), $\delta =1$ (middle), and $\delta=5$ (right).}
    \label{fig:mnist_delta}
\end{figure*}

\smallskip\noindent\textbf{The impact of $\delta$.}
Second, we explore the impact of the value of $\delta$ on robustness.
We ran our experiments on the SCC aggregator, where $\delta$ is a hyper-parameter fixed during training that we can directly control. 
We let $\delta$ take the values 0.5, 1, and 5.

We report in Figure~\ref{fig:mnist_delta} the accuracy of users' models during benign training (top), and the distance of users' models to the adversarially chosen target model (the all-0 model) under the state-override attack (bottom), for various values of $\delta$.
As expected by our theoretical findings, both the speed at which the adversary can manipulate the model and the speed at which honest users learn are proportional to the value of $\delta$.
The larger $\delta$, the faster users benefit from collaboration but the faster the adversary overwrites the users' models.
When $\delta = 0.5$ (left), there is little space for manipulation, and the distance to the target model decreases very slowly (bottom left). 
The attack would need much more epochs to take full control of users' models.
At the same time, a small $\delta$ value reflects on the quality of learning, which is null (accuracy does not increase in the benign scenario in the given number of epochs, top left).
The small value $\delta = 0.5$ hinders both the adversary and users towards their respective tasks without distinction, i.e., manipulation and learning.
For $\delta=1$ (middle), the learning capabilities of users improve during benign training (top middle) and the adversary increases its influence on the users model during the stat-override attack (bottom middle).
Although in 50 epochs, neither the honest users have reached optimal performance nor the adversary has taken over the model. Yet, the influence of the adversary over users' models is definitely sufficient to control one or few of their weights (for instance to insert a backdoor).
When $\delta=5$ (right), learning is fast among honest users (top right). But, the adversary eventually gets full control over all users' model, reaching an L2 distance to the target close to 0 just under 50 epochs (bottom right), i.e., $\Delta(m_{target}, m_{users})$ goes close to $0$.


\smallskip\noindent\textbf{The case of dynamic $\delta$.}
We now explore the setup where $\delta$ is not fixed throughout training but dynamically computed at each epoch and by each user, using network inputs. 
Again, we ran our experiments on the SCC aggregator, where such a choice of dynamic $\delta$ is proposed and used in experiments.
At epoch $t$, each user $i$ computes the robustness threshold $\delta$ as the variance among neighbors, i.e.:

\begin{equation}
\label{eq:dynamic}
    \delta_i^t := \sqrt{\sum_{j}\frac{1}{n}||\theta_i^{t}-\theta_j^{t}||_2^2} \quad.
\end{equation}

Since all neighbors contribute to the computation of this dynamic threshold, we find that the computation of the threshold can now be manipulated by an adversary in the network.
Through purposefully increasing the variance among users, e.g., by introducing dissensus into the network~\cite{scclipping}; the adversary can in turn increase the value of $\delta_i^{t}$, facilitating any following poisoning attack. 

\begin{figure}[!t]
    \centering
    \includegraphics[width=.41\textwidth]{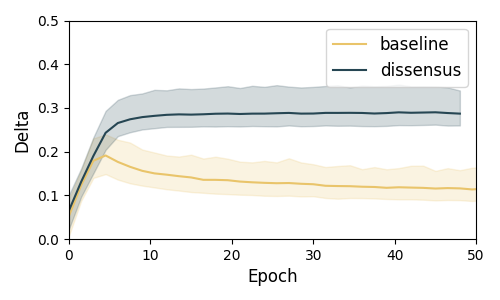}
    \caption{Evolution of $\delta$ during benign training (baseline) and under the dissensus attack.}
    \label{fig:tau}
\end{figure}

In Figure~\ref{fig:tau} we show the result of using the \textit{dissensus attack}~\cite{scclipping} in which the adversary $\adv$ sends to their neighbors the gradient that maximizes distance between updates, i.e.: 

$$\theta_{\adv i}^t = \underset{j}{\sum} \frac{1}{n}*(\theta_i^0-\theta_j^0).$$

This simple attack is enough to almost triple the average value of $\delta_i^t$. By increasing three-fold the value of threshold, the adversary triples the speed of its attack.
We conclude that using a dynamic $\delta$ computed on network inputs actually increases the attack surface.

\subsubsection{Impact on real systems}\

\noindent We now validate our results further in two use cases of CML proposed in the literature in which robustness is crucial.
We reproduce their setup, and show that an adversary implementing the state-override attack (with target model the all-0 model) can easily push the final model to misbehave, leading to very concrete potential harms.

\smallskip\noindent\textbf{Collaborative Learning for Health.}
There is an increasing number of publications applying CML to medical data, showing that it constitutes a realistic use case which has been implemented in real-world scenarios~\cite{health10}.
We take as a representative example the work of Grama et al.~\cite{robFLHealth}.
In this work, the learning task is to predict diabetes from medical data, and they use the Pima Indians Diabetes dataset~\cite{diabetes}.

As in the original paper, we use a shallow 3-layers CNN, combining the layers by ReLu; and default hyper-parameters: an Adam optimizer with a learning rate of $1e-5$, batch size of 10, and cross entropy as loss function.
We consider 10 users, out of which 2 are malicious. We distribute the data among them in the following way: three of them own 39 samples, three own 59 and four own 80.
We split the data among users in an IID fashion.
Following the experimental section of Grama et al.~\cite{robFLHealth}, we use the error rate as the success metric on the learning task. 

\begin{figure}[!tbp]
    \centering
    \includegraphics[width=.42\textwidth]{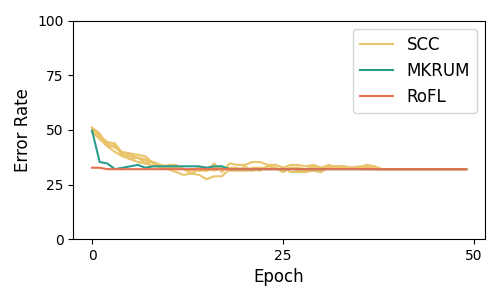}
    \caption{Error Rate under the state-override attack on the diabetes task.}
    \label{fig:er_health}
\end{figure}

\begin{figure*}[!htbp]
    \centering
    \includegraphics[width=\textwidth]{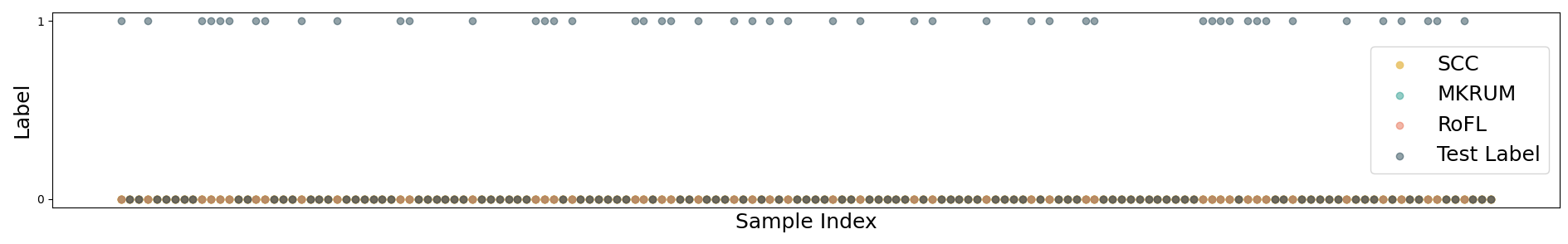}
    \caption{Predictions of final models under the state-override attack on the diabetes task.}
    \label{fig:pred_health}
\end{figure*}

We show the error rate of honest users' models for the three robust aggregators SCC, MKRUM, and RoFL in Figure~\ref{fig:er_health}.
Under attack, the error rate of these models converges to 32\%.
This error rate coincides with the best rate achieved in the original work. This could be interpreted as a success of robust aggregators and could mislead us into thinking that the attack is not successful.
However, when plotting the actual predictions made by the final models against the expected ground truth labels (see Figure~\ref{fig:pred_health}), we can see the impact of the attack: for all three aggregators, the honest users' final models exclusively predict `0' labels (`No diabetes' diagnosis).
The `low' error rate of 32\% is then just a consequence of the fact that the majority of users in the test dataset do not have diabetes.
If this system were to be used in practice, it would be very damaging, as it would never detect diabetes in patients.

\begin{figure}[!htbp]
    \centering
    \includegraphics[width=.45\textwidth]{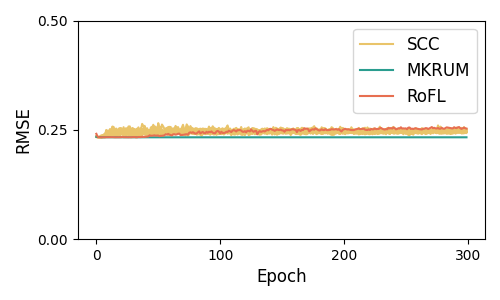}
    \caption{RMSE under the state-override attack on the driving task.}
    \label{fig:rmse}
\end{figure}

\begin{figure*}[!htbp]
    \centering
    \includegraphics[width=\textwidth]{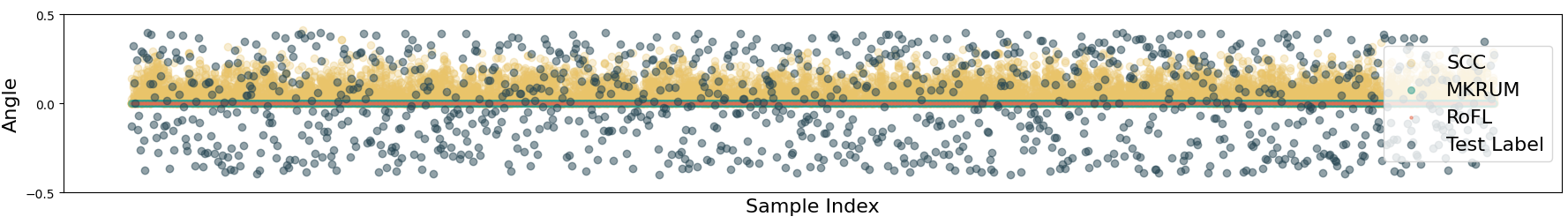}
    \caption{Predictions of final models under the state-override attack on the driving task.}
    \label{fig:pred_driving}
\end{figure*}

\smallskip\noindent\textbf{Collaborative Learning for Autonomous Driving.}
We also observe that reliable autonomous driving is a common application of CML. We select the work of Nguyen et al.~\cite{FLdeepdriving} as a representative example. 
In this work, the learning task is to predict the steering angle required for following the road ahead, based on a photograph of the road in front of a car.

Following the original paper, we consider 11 users (vehicles) and use the udacity dataset~\cite{udacity} for the task.  
We divide the dataset uniformly among users.
We use the FADNet model architecture~\cite{FLdeepdriving}, a CNN architecture based on ResNet18, the RMSE as loss function, a batch size of 32, a learning rate of 0.001, and the Adam optimizer.
We use only 300 training epochs instead of 3000 in the original paper.
Similarly to the experimental section of Nguyen et al.~\cite{FLdeepdriving}, we use RMSE as the performance metric.

Since no threat model is provided in the original work, we consider a weak adversarial model with only one of the users being adversarial.

We report the RMSE of honest users' models across training in Figure~\ref{fig:rmse}.
We see that the error stalls at around 0.25 for all aggregators.
While the error is rather steady and does not increase, this value is double the RMSE of the final model reported in the original work (0.107).
To show further the impact of the attack in this task, we report in Figure~\ref{fig:pred_driving} the predictions of the honest users' final model, and compare them to the ground truth.
We see that using MKRUM and RoFL as robust aggregators results in training a model which produces only `0°' predictions.
Since these algorithms correspond to a federated setup, all honest users share the same global model.
Because of the resulting visual overlap, predictions made with the model trained with RoFL (orange) are plotted using slightly smaller dots to make visible the predictions of the final model trained using MKRUM (light blue) that would be hidden below otherwise.
For SCC, we report the performances of all honest users' local models, which are all only positive and much smaller than the ground truth steering angle.
So, as in the health application, in this task and setup, the adversary can successfully manipulate the honest users' models for all tested robust aggregators.
In practice, using the resulting models would mean autonomous cars turning less than needed or not turning at all, threatening the life of passengers and bystanders.



\subsection{Takeaways}
In this section, we show that, unless the participant has enough knowledge about the learning task to choose $\Delta$, $\delta$, and $\theta^*$ such that $R$ accurately distinguishes malicious updates, a strategic adversary can always successfully manipulate honest users' models.
However, we demonstrate that a participant with aforementioned knowledge about the learning task does not benefit from collaboration, and thus has no incentive to participate.

We furthermore show that without this guarantee, distance-based robust aggregators are not useful towards robustness beyond limiting the magnitude of the contributions of \textit{both} honest and malicious users at each epoch through the use of a threshold.
As consequence, there is no inherently \textit{good} value of $\delta$ which allows learning and robustness of the final model at the same time.

We highlight this trade-off in current state-of-the-art distance-based robust aggregators, and further show that they do not offer any guarantee and followingly fail to accurately allow participants to accurately filter out adversarial updates.
A strategic adversary can successfully influence the predictions of honest users' models for two state-of-the-art distance-based robust aggregators and for a variety of tasks, including two real-life applications.
While our results cover a broad number of cases but not all of them, we argue that there either currently exists or one can design an attack that will support our theoretical findings against other distance-based aggregators. 

In other words, we demonstrate that \textbf{for any distance metric, robustness can only be achieved at the cost of removing learning gains}. Whenever there is no room for manipulation, there is also no room for learning.


\section{Evaluating Behavior is as Hard as Learning}

We now analyze behavior-based aggregators, i.e., those where participants use evaluation functions of the form $R(\theta, \mathcal{D}) \doteq \pi(\theta; \mathcal{D}) - \pi(\theta^*; \mathcal{D})$.
As we are in the CML setting, participants only have partial knowledge of the input domain $\mathcal{D}$ through their local dataset $S$.
Thus, they cannot \textit{exactly} compute $R(\theta, \mathcal{D})$, and can only have an approximation based on their local data: $R(\theta, S)$.
We explore how such an approximation impacts security and how it relates to the ability of participants to learn from their neighbors.


\subsection{The Implications of Learning}

We first instantiate the oracles in the R-IND game such that this game assesses the capability of the distinguisher to estimate the behavior-based evaluator $R$ locally, rather than the effectiveness of the evaluator to distinguish malicious updates. We achieve this by instantiating the adversarial oracle $\mathcal{O}_\texttt{malicious}$ to produce updates $\theta$ such that $R(\theta, \mathcal{D}) > \delta$; and the benign oracle $\mathcal{O}_\texttt{benign}$ to produce updates $\theta$ such that $R(\theta, \mathcal{D}) \leq \delta$.

\begin{theorem}~\label{theorem:behavior}
In such an instantiation, when $R(\theta, \mathcal{D})$ is behavior-based, the advantage $Adv_\userP^\text{R-IND}(\delta; S_\userP, \mathcal{D})$ of the participant $\userP$ playing the R-IND($\delta, S_\userP, \mathcal{D})$ game is proportional to its advantage $Adv_\userP^\text{L-EUF}(\epsilon; S_\userP, \mathcal{D})$ in playing the L-EUF($\epsilon, S_\userP, \mathcal{D})$ game.
\end{theorem}

The implication of Theorem~\ref{theorem:behavior} is that a participant capable of accurately evaluating received updates does not have much potential to learn from others. 
Conversely, a participant with large learning potential is expected to make mistakes when evaluating received updates.

\begin{proof}[Proof Sketch]
We start by looking at the R-IND game instantiated with the oracles above, and a participant (distinguisher) $\userD$ playing it.
The distinguisher, with local dataset $S_\userD$, uses $R(\theta, S_\userD)$, an approximation of $R(\theta, \mathcal{D})$, to evaluate updates they receive.

When the behavior of $\theta$ (or of the model resulting on the use of the update), and $\theta^*$ respectively, on samples in $S_\userD$ is close to the behavior on samples in $\mathcal{D}$, the approximation $R(\theta, S_\userD)$ is good, and $\userD$ can compare it to $\delta$ to successfully distinguish benign updates from malicious ones, and win the game.
When the behavior of $\theta$ on samples in $S_\userD$ is very different from its behavior on samples in $\mathcal{D}$, $R(\theta, S_\userD)$ is a bad estimation of $R(\theta, \mathcal{D})$ and comparing this value to $\delta$ to classify updates will result on many errors.

This highlights that the advantage of $\userD$ in the R-IND game is a function of how well $S_\userD$ represents $\mathcal{D}$, according to $\pi$.
Let's say that the local dataset $S_\userD$ is $\epsilon-$representative of $\mathcal{D}$~\cite[Definition 4.1]{ltheory}, for a given $\epsilon$.
According to the Lemma 4.2 of~\cite{ltheory}, $\userD$ can learn a model $m$ such that $\underset{(x, y)\sim\mathcal{D}}{\mathbb{E}}[\mathcal{L}(m, (x, y))] - \underset{m^*}{\text{ min }}\underset{(x, y)\sim\mathcal{D}}{\mathbb{E}}[\mathcal{L}(m^*, (x, y))] < 2\epsilon$ and win the L-EUF$(2\epsilon; S_\userD, \mathcal{D})$ game. 
So, how well $S_\userD$ represents $\mathcal{D}$ also impacts the ability of $\userD$ to win the L-EUF game. 
If the participant can win the L-EUF game without requiring collaboration of other users, its learning potential is negligible.


%


We now look at the L-EUF game, played by a learner $\userL$.
Let us consider a dataset $S_i$ such that $Adv_\userL^\text{L-EUF}(\epsilon; S_\userL \cap S_i, \mathcal{D}) - Adv_\userL^\text{L-EUF}(\epsilon; S_\userL, \mathcal{D})$ is non-negligible, i.e, $\userL$ can significantly learn from $S_i$.
Then, there exist a part of $\mathcal{D}$, captured by $S_i$, that $S_\userL$ cannot represent, i.e., that a model trained on $S_\userL$ cannot generalize to, and that they cannot win the L-EUF game using $S_\userL$.
This also means that $R(\theta, S_\userL)$ is a bad approximation of $R(\theta, \mathcal{D})$ on samples in $S_i$, and in this subset of $\mathcal{D}$ $\userL$ cannot accurately distinguish malicious updates from benign ones. Thus, they cannot win the R-IND game.
The larger the gap between $S_\userL$ and $S_i$, the more learning potential $\userL$ has, but smaller their advantage in winning the R-IND game due to the poor approximation of $R(\theta, \mathcal{D})$.
\end{proof}

\subsection{Validation of the Learning versus Quality of Robustness trade-off}

\begin{figure*}[!ht]
    \centering
    \begin{subfigure}{0.45\textwidth}
    \includegraphics[width=\textwidth]{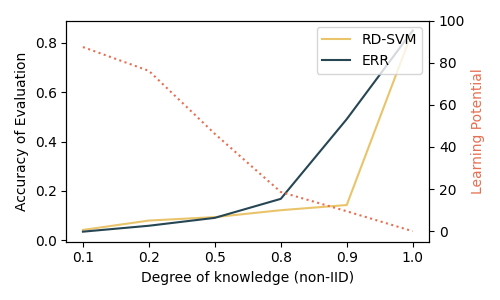}
    \end{subfigure}
    \begin{subfigure}{0.45\textwidth}
    \includegraphics[width=\textwidth]{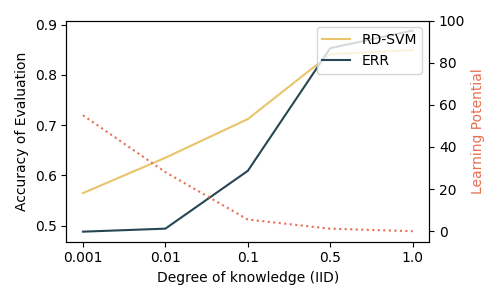}
    \end{subfigure}

    \begin{subfigure}{0.45\textwidth}
    \includegraphics[width=\textwidth]{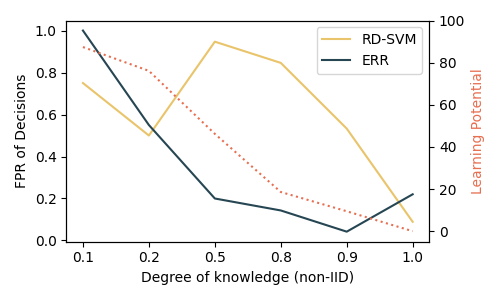}
    \end{subfigure}
    \begin{subfigure}{0.45\textwidth}
    \includegraphics[width=\textwidth]{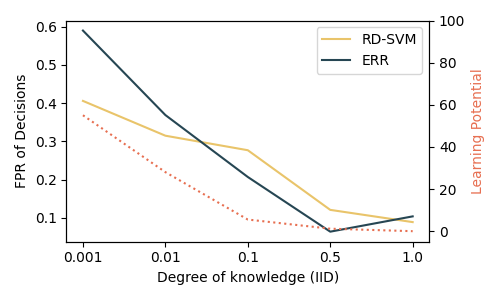}
    \end{subfigure}
    \caption{Accuracy (top) and FPR (bottom) of the decisions of the test user versus its learning potential for increasing knowledge (non-IID, left; IID, right).}
    \label{fig:acc_learning}
\end{figure*}

To validate the implications of Theorem~\ref{theorem:behavior}, we empirically investigate the relationship between learning potential and the quality of evaluation made by participants when using two behavior-based aggregators from the literature: RD-SVM~\cite{svmdl} and ERR~\cite{local} (see Section~\ref{sec:examples-behavior}).
We study this relation under different degrees of knowledge by varying both the quality and the quantity of the participants' local training data (since both data representativeness and volume are needed to succeed at the learning task).

\subsubsection{Results on MNIST}\

\smallskip\noindent\textbf{Setup.}
We consider a network of 10 participants. 
We use the MNIST dataset, which we divide in train and validation sets; a shallow 3-layers CNN model architecture; a default learning rate of 0.01; a batch size of 32; and an Adam optimizer. 
Out of the 10 participants, we let 9 users be used as control subjects.  
Each of these users is given a partition of the MNIST train dataset, ranging from 5 to 15\% of the available data, split in an IID fashion. 
The remaining participant is our test subject, whose local dataset simulates different levels of knowledge on the learning task at hand. 
We run experiments in two scenarios.
One in a peer-to-peer scenario where the test subject doing the robustness evaluations is a user, 
The other in a federated scenario where the test subject doing the robustness evaluations is the server.
In each scenario we run two sets of experiments. In our first set of experiments, we give the test subject an increasing amount of knowledge which grows in a non-IID manner to simulate an increasing knowledge with respect to \emph{quality} (representativeness): the test subject starts only knowing elements of the MNIST dataset with label `0', then `0' and `1', then `0', `1', and `2', \ldots until they know all the labels, hence the whole training dataset. 
In our second set of experiments, we give the test subject an increasing amount of knowledge which grows in an IID manner, simulating an increasing knowledge with respect to \emph{quantity} (volume): in the first training process we assign to the subject a dataset which consists in 0.1\% of the MNIST train dataset, sampled uniformly across all labels, in the second training process, the dataset grows to 1\% of the MNIST train dataset, then 10\%,\ldots until the test subject has knowledge of the whole training dataset.

Finally, instead of constructing the $\mathcal{O}_\texttt{benign}$ and $\mathcal{O}_\texttt{malicious}$ oracles, we directly use the updates produced by the users during the training processes, and use $R(\theta, \mathcal{D})$ as the ground truth for their benignness, with $\mathcal{D}$ the whole MNIST dataset (including validation data). 
When $R(\theta, \mathcal{D}) < \delta$, the update $\theta$ is considered benign, and otherwise malicious.




\smallskip\noindent\textbf{Results in the peer-to-peer setup.} 
In Figure~\ref{fig:acc_learning}, we plot the evolution of the performance of the decisions of the test user for the RD-SVM and ERR robust aggregators. 
We plot both the evolution of accuracy (top) and false positive rate (bottom), for data increasing in quality (left) and in quantity (right).
Continuous lines represent the test user's success in distinguishing benign updates for different degrees of knowledge. 
The dashed line represents the test user's learning potential, i.e., the difference in accuracy between a model trained only with the test user's data and the accuracy of a model trained with the data of all users participating in the collaborative learning training.

We first look at the evolution of the accuracy of the test user' evaluations (top).
As the test user's knowledge increases, i.e, as they learn more labels or have access to more data, their learning potential diminishes.
Indeed, with increasing knowledge, the performance of a model trained with their local data only improves (accuracy increases) and consequently, their potential to learn from others decreases.
At the same time, the accuracy of the decisions of the test user increases. 
When presented with an update, the test user can more accurately evaluate its benignness compared to the ground truth as their knowledge increases.
\textbf{As stated by Theorem~\ref{theorem:behavior} and its implication, the accuracy of robustness evaluation of the test user is inversely proportional to its learning potential for both evaluators.}

We observe that the accuracy in the IID case (top right) has a much smoother evolution than in the non-IID setting (top left), where accuracy grows significantly when the user has data from 8 labels. This reinforces our hypothesis that it is hard to generalize on labels that are not known.
It is also worthy to note that the subject never reaches an accuracy of 1, even when they have maximal knowledge (degree 1, representing all data available in the network).
This is because, collectively, the users only know of the MNIST \textit{train} dataset, while $\mathcal{D}$ also contains a validation set.
Thus, even if the test user's local data $S_\userD$ is the union of the data of all users, $R(\theta, S_\userD)$ remains an imperfect approximation of $R(\theta, \mathcal{D})$.

Finally, we study the False Positive Rate (FPR, bottom), since false positives capture malicious updates classified as benign, which is the most harmful event.
The FPR is proportional to learning potential for both evaluators.
Furthermore, when the test user has little knowledge, which is the expected situation in realistic cases, their FPR is rather high.
For instance, when the test user knows 10\% of the dataset in an IID fashion--- which corresponds to a setup where users have uniform amount of data, i.e., $\frac{100}{\text{number of users}}\%$ of MNIST dataset---, the FPR is of respectively 0.315 and 0.369 for the RD-SVM and ERR algorithms.
This means that among all malicious updates that the user received, it wrongly classifies as benign around a third of them.
In the non-IID setting (bottom left), when the user knows two labels, half of the updates classified as benign are actually detrimental to the model. 
These false positive rates are extremely high, especially for a scenario \emph{where all participants are honest}. A malicious user crafting updates to evade detection would have a much higher success rate, creating even more damage.

\begin{figure*}[!ht]
    \centering
    \begin{subfigure}{0.45\textwidth}
    \includegraphics[width=\textwidth]{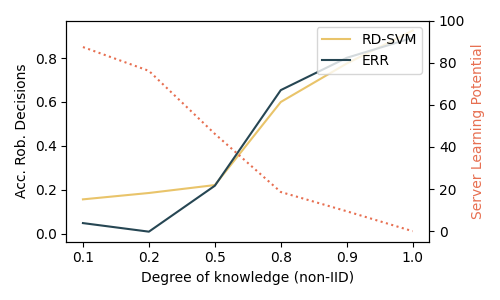}
    \end{subfigure}
    \begin{subfigure}{0.45\textwidth}
    \includegraphics[width=\textwidth]{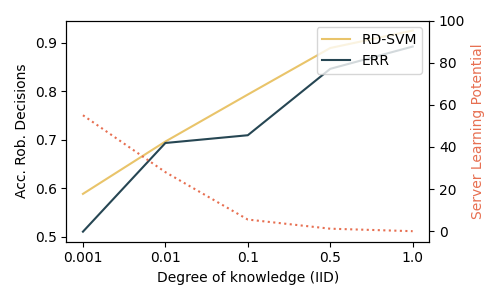}
    \end{subfigure}

    \begin{subfigure}{0.45\textwidth}
    \includegraphics[width=\textwidth]{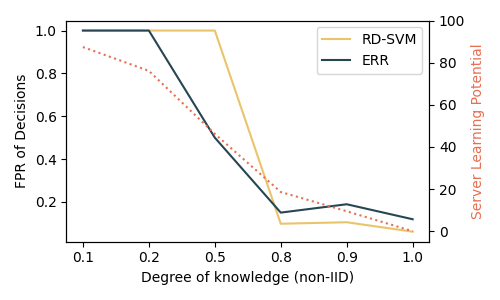}
    \end{subfigure}
    \begin{subfigure}{0.45\textwidth}
    \includegraphics[width=\textwidth]{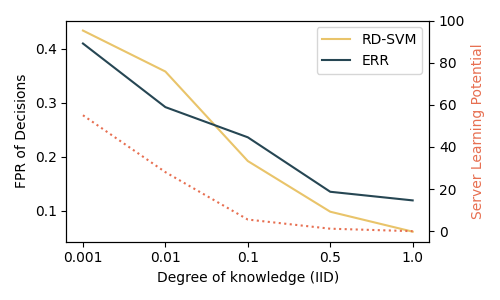}
    \end{subfigure}
    
    \caption{Accuracy (top) and FPR (bottom) of the decisions of the server versus its learning potential for increasing knowledge (non-IID, left; IID, right).}
    \label{fig:server_acc_learning}
\end{figure*}

\smallskip\noindent\textbf{Results in the federated setup.} 
In Figure~\ref{fig:server_acc_learning}, we report the evolution of the accuracy and FPR of the decisions of the server against the server learning potential for RD-SVM and ERR. 
The results are similar to the peer-to-peer case: the server is subject to the fundamental trade-off stated in Theorem~\ref{theorem:behavior}.
Whether the server evaluates updates based on a test set obtained from the users --at the cost of privacy-- or from public data, the quality of the server decisions depend on the quality of this test set. 
If this test set is not representative of $\mathcal{D}$ (left most part of the graphs), it leads to poor evaluation from the server of received updates.
If the test set enables the server to successfully distinguish malicious updates (right most part of the graphs), then the server learning potential is at its lowest.
In other words, at that point the server does not benefit from the users' inputs from the users with respect to the learning task, and collaboration does not provide better learning over a strategy where the users would directly use a model locally computed by the server using $S_\userD$.

\subsection{Takeaways}
In this section, we provided a theoretical analysis of behavior-based evaluators that reveals that, for both a server in the federated setting, and a user in decentralized learning, \textbf{the ability to distinguish between benign and malicious updates depends on their local knowledge and is proportional to their potential to get benefits from participating in the collaborative learning process}.
If the local data of participants is representative of the true distribution and quantitatively enough, they are able to make accurate evaluations. At that point, however, participants can also learn an optimal model for the task on their own, limiting the benefits of collaboration. If the data they hold is not representative, then they get a lot of benefits, but robustness cannot be guaranteed.


\section{Related Work}

There is a large number of publications about poisoning attacks in CML, e.g.,~\cite{localpoison, alie, imp, history, backdoor, scclipping, bylan, limits}, but as opposed to our work, their results are hard to generalize to families of robust aggregators.
Among existing attacks, a particularly relevant work is the one of Ozfatura et al.~\cite{history}.
The authors show that centered clipping~\cite{ogclipping}, a distance-based robust aggregator whose robustness assumption relies on honest users' models having a small variance, does not eliminate the attack surface completely.
They design a strategic adversary who chooses the best adversarial input to the aggregator under a norm constraint and eventually successfully manipulates the honest users' models. 
This work confirms our findings on distance-based aggregators by providing one concrete example of how to exploit the specific distance-based aggregator they study.
Ozfatura et al. propose an improvement to the aggregator which consists in randomly altering the reference point, i.e., $\theta^*$. 
However, as the authors do not evaluate the flaws of the design principles of the attacked aggregator and its family, the fix they propose defends against their attack but does not solve the problem pointed out in this paper, giving a false sense of security. 
Given Theorem~\ref{theorem:dist}, if the fix allows for learning, then there exists a strategic adversary who would be able to manipulate the training even against this fix, as in the other examples we evaluate in Section~\ref{sec:eval-distance}. 
Looking at this problem using our framework would have shown that minimal fixes would not cut short the arms race by directly showing the fundamental trade-off associated with the choice of evaluator.

A related work that showcases the limitations of robust aggregators is the work of Khan et al.~\cite{pitfalls}, but they do so from an angle different from this paper.
Through a series of experiments, the authors show how hidden assumptions in the evaluation setup of robust aggregators can misrepresent robustness and give a false sense of security. 
For instance, the data distribution among participants can affect the validity of robustness claims: some robust aggregators only work when the data is distributed in an IID fashion (which means that users have qualitatively good local knowledge).
Their evaluation can be seen as an empirical validation of our formal results, e.g., Theorem~\ref{theorem:behavior}.

\section{Conclusion}

In this work, we provide the first systematization of robustness evaluators in Collaborative Machine Learning; and the first formalization of the robustness goal in presence of a malicious adversary. Using our formalization, we analyse the two prevalent robust aggregator families in the literature: distance-based and behavior-based.

Firstly, we find that when the evaluation is distance-based, regardless of the distance metric and the threshold chosen by the defender, an adversary can build malicious updates that the distance-based evaluation will fail to identify.
We show how this vulnerability can lead to catastrophic effects in a health and a self-driving application.
We conclude that distance is not a good proxy for maliciousness and that distance-based definitions of $R(\theta, \mathcal{D})$ are inadequate to implement robust aggregators in collaborative learning.

Secondly, we find that users can only evaluate updates based on the model behavior when the user's local knowledge is sufficiently large. At that point, however, this user can learn a very limited amount of information from other users and benefits very little from participating in collaborative learning, limiting the advantages to other criteria such as a hypothetical gain in efficiency in the training process.

Overall, we show that (a) \textbf{contrary to the common belief of the CML community, state-of-the-art robust aggregators do not provide robustness against a strategic adversary}; and (b) \textbf{the notions of robustness implemented by current robust aggregators is directly in conflict with learning}.
As this field continues to grow, we hope that our work can help redirect the efforts of the community, avoiding directions in which the arms-race can not conclude favorably.


\smallskip\noindent\textbf{Future Work.}
Although the two classes we study encompass all techniques used by robust aggregators proposed in the literature, we have not evaluated hybrid approaches, e.g. aggregators which combine both distance- and behavior-based evaluation techniques.
We do not expect these aggregators to escape the learning/robustness trade-off, since distance-based evaluation does not help with the flaws of behavior-based evaluators and vice-versa.
It might still be valuable to use our games and reasoning to formally analyze their limitations.

Last, while we demonstrate that existing evaluation functions do not enable robust distinguishing of malicious updates, there may be trade-offs between security and learning acceptable for certain use-cases. For instance, not in health-oriented scenarios, but when CML is used for advertising-related predictions. Thus, it would be interesting to study how our framework could be used to find  application-specific ``good'' values for parameters and functions such as $\delta$, $\Delta$, or $\pi$.



\newpage

\bibliographystyle{IEEEtranS}
\bibliography{IEEEabrv, bib}
%

\end{document}